\documentclass[11pt,a4paper]{article}

\usepackage[utf8]{inputenc}
\usepackage[T1]{fontenc}
\usepackage{lmodern}
\usepackage[margin=2.5cm]{geometry}
\usepackage{amsmath,amssymb,amsfonts}
\usepackage{booktabs}
\usepackage{multirow}
\usepackage{graphicx}
\usepackage{xcolor}
\usepackage{hyperref}
\usepackage{cleveref}
\usepackage{caption}
\usepackage{subcaption}
\usepackage{enumitem}
\usepackage{microtype}
\usepackage{float}
\usepackage{natbib}
\usepackage{setspace}
\usepackage{titlesec}
\usepackage{lineno}

\hypersetup{
    colorlinks=true,
    linkcolor=blue!60!black,
    citecolor=green!50!black,
    urlcolor=blue!70!black,
}

\titlespacing*{\section}{0pt}{1.5ex plus 0.5ex}{1ex plus 0.3ex}
\titlespacing*{\subsection}{0pt}{1.2ex plus 0.4ex}{0.8ex plus 0.2ex}
\titlespacing*{\subsubsection}{0pt}{1ex plus 0.3ex}{0.6ex plus 0.2ex}

\newcommand{\eg}{\textit{e.g.}}
\newcommand{\etal}{\textit{et al.}}

\newcommand{\tok}[1]{\texttt{#1}}

\title{%
    \textbf{Sparse Autoencoder Decomposition of Clinical Sequence Model\\ Representations: Feature Complexity, Task Specialisation,\\ and Mortality Prediction}
}

\author{
    Chris Sainsbury\textsuperscript{1,2,3}
    \and
    Feng Dong\textsuperscript{5}
    \and
    Andreas Karwath\textsuperscript{4}
    \\[1ex]
    \textsuperscript{1}School of Cardiovascular \& Metabolic Health, University of Glasgow, UK \\
    \textsuperscript{2}School of Medicine, University of Dundee, UK \\
    \textsuperscript{3}NHS Greater Glasgow and Clyde, UK \\
    \textsuperscript{4}Cancer and Genomic Sciences, School of Medical Sciences,\\ College of Medicine and Health, University of Birmingham, UK \\
    \textsuperscript{5}Department of Computer and Information Sciences, University of Strathclyde, UK \\[1ex]
    \texttt{chris.sainsbury@glasgow.ac.uk}
}

\date{}

\begin{document}
\maketitle

% ═════════════════════════════════════════════════════════════════════════════
% ── Abstract ──────────────────────────────────────────────────────────────
% ═════════════════════════════════════════════════════════════════════════════

\begin{abstract}

Sparse autoencoders (SAEs) have been applied to large language models and protein language models, but not systematically to electronic health record (EHR) foundation models.  We train TopK SAEs on FlatASCEND, a 14.5-million-parameter autoregressive clinical sequence model, at all 10 residual stream extraction points on INSPECT (outpatient) and MIMIC-IV (ICU).  SAE decomposition reveals progressive abstraction across transformer depth: layer-0 features are near-perfect token detectors (45.7\% singleton), while layer-6 features span approximately 30 token types across multiple clinical categories (0.5\% singleton).  Under full-sequence simple linear probes, SAE features outperform dense representations for discrete event prediction (mortality) while dense representations outperform for continuous magnitude prediction (length of stay)---a probe-level representational phenomenon that does not extend to clinically relevant leakage-safe windows, where dense representations match or exceed SAE features across all tested settings (eICU-CRD 48-hour AUC: SAE 0.871 versus dense 0.880; base model zero-shot, SAE dictionaries trained on eICU activations; MIMIC-IV: 0.836 versus 0.914; INSPECT 1-year/3-year: 0.697 versus 0.800).  A delta-mode intervention method reduces SAE perturbation noise by 86$\times$, enabling cleaner feature-level experiments, though the resulting perturbation effects are larger than random controls in 3 of 4 conditions but not formally significant.  Feature reproducibility across random seeds is 21\%, and individual features should be interpreted as illustrative rather than stable.

\end{abstract}

% ═════════════════════════════════════════════════════════════════════════════
\section{Introduction}
\label{sec:introduction}
% ═════════════════════════════════════════════════════════════════════════════

Transformer-based foundation models trained on electronic health records (EHRs) are increasingly effective at downstream clinical prediction tasks,\citep{li2020behrt,rasmy2021medbert,guo2022clmbr,steinberg2023motor} with recent models achieving zero-shot disease forecasting AUC exceeding 0.79 for heart failure.\citep{rajamohan2026raven,chen2025nep}  These models learn distributed representations of clinical data, but what these representations encode---and how clinical concepts emerge through transformer depth---remains largely unexplored.

Existing interpretability methods for EHR models are predominantly post-hoc.  SHAP values\citep{lundberg2017shap} and attention visualisations attribute predictions to input features but do not decompose the model's internal representations into meaningful units.  They answer ``which inputs mattered'' but not ``what concepts has the model learned.''  Sparse autoencoders (SAEs) offer a different approach: by training an overcomplete dictionary on a model's internal activations, SAEs decompose distributed representations into sparse linear combinations of interpretable features.\citep{bricken2023monosemanticity,cunningham2023sae}  This methodology has been applied to large language models,\citep{templeton2024scaling} protein language models,\citep{orlov2026review} and pathology vision transformers,\citep{le2024pluto} but SAE analysis of EHR foundation models has not, to our knowledge, been systematically examined.  Concurrent work by Modi \etal\citep{modi2026medsae} applies SAEs to medical LLMs (MedGemma-27B, OpenBioLLM-70B) processing clinical discharge note text, revealing architecture-dependent polysemy encoding and distributed medical knowledge.  Our work complements this by applying SAEs to an EHR foundation model operating on coded clinical sequences rather than free text, revealing task-dependent representational specialisation not previously reported.

We apply TopK SAEs\citep{gao2024topk} to FlatASCEND, a 14.5-million-parameter autoregressive clinical sequence model that generates multi-step clinical trajectories with continuous inter-event time prediction.\citep{sainsbury2025flatascend}  In a companion paper, we showed that FlatASCEND's generated trajectories partially recover directionally consistent pharmacological associations (4 of 10 correct mechanistic drug--outcome directions, plus 2 treatment-context associations, on MIMIC-IV at the patient level).\citep{sainsbury2025flatascend}  Here we decompose the model's internal representations to characterise what clinical concepts emerge, how they specialise for different prediction tasks, and whether they support mortality prediction at leakage-safe observation windows.

The specific contributions are:

\begin{enumerate}[leftmargin=2em]
    \item \textbf{SAE analysis of an EHR foundation model}: TopK SAEs trained at all 10 residual stream points on INSPECT and MIMIC-IV, with task-dependent evaluation extended to four datasets including eICU-CRD zero-shot transfer, with feature complexity increasing ten-fold from token detectors at layer 0 to distributed clinical concepts at layer 6 (Section~\ref{sec:sae-results}).

    \item \textbf{Task-dependent representational specialisation}: under full-sequence simple linear probes, SAE features outperform dense representations for discrete event prediction (mortality) while dense representations outperform for continuous magnitude prediction (length of stay), a pattern that persists at $K$ = 128; at leakage-safe windows, this advantage is reduced or reversed (Sections~\ref{sec:specialisation}--\ref{sec:mortality}).

    \item \textbf{Leakage-safe mortality prediction}: setting-dependent observation windows (48 hours ICU, 365 days outpatient) as an approach to evaluating clinical prediction without end-of-stay information leakage; at leakage-safe windows, dense representations matched or exceeded SAE features across all datasets tested, with a small gap on eICU-CRD (0.871 versus 0.880) and a larger gap on MIMIC-IV (0.836 versus 0.914) (Section~\ref{sec:mortality}).

    \item \textbf{Delta-mode intervention}: a noise-reduced method for SAE-based feature perturbation that reduces the intervention noise floor by approximately 86$\times$ relative to na\"{i}ve reconstruction (Section~\ref{sec:perturbation}).

    \item \textbf{Reproducibility characterisation}: feature stability across random seeds is 21\%; the core qualitative conclusions appeared stable in the runs examined, but downstream seed-wise variability was not formally quantified (Section~\ref{sec:reproducibility}).
\end{enumerate}

% ═════════════════════════════════════════════════════════════════════════════
\section{Results}
\label{sec:results}
% ═════════════════════════════════════════════════════════════════════════════

\subsection{SAE decomposition reveals progressive abstraction across transformer depth}
\label{sec:sae-results}

We trained TopK sparse autoencoders\citep{gao2024topk} on FlatASCEND's residual stream at all 10 extraction points (layers 0--9), decomposing 384-dimensional hidden states into 3,072 sparse features (8$\times$ expansion, $K$ = 16 active features per input).  SAEs were trained independently per layer per dataset using streaming extraction with dead feature resampling.

\subsubsection{Layer-specialised representations: the explained variance U-curve}

\begin{figure}[t]
\centering
\includegraphics[width=\textwidth]{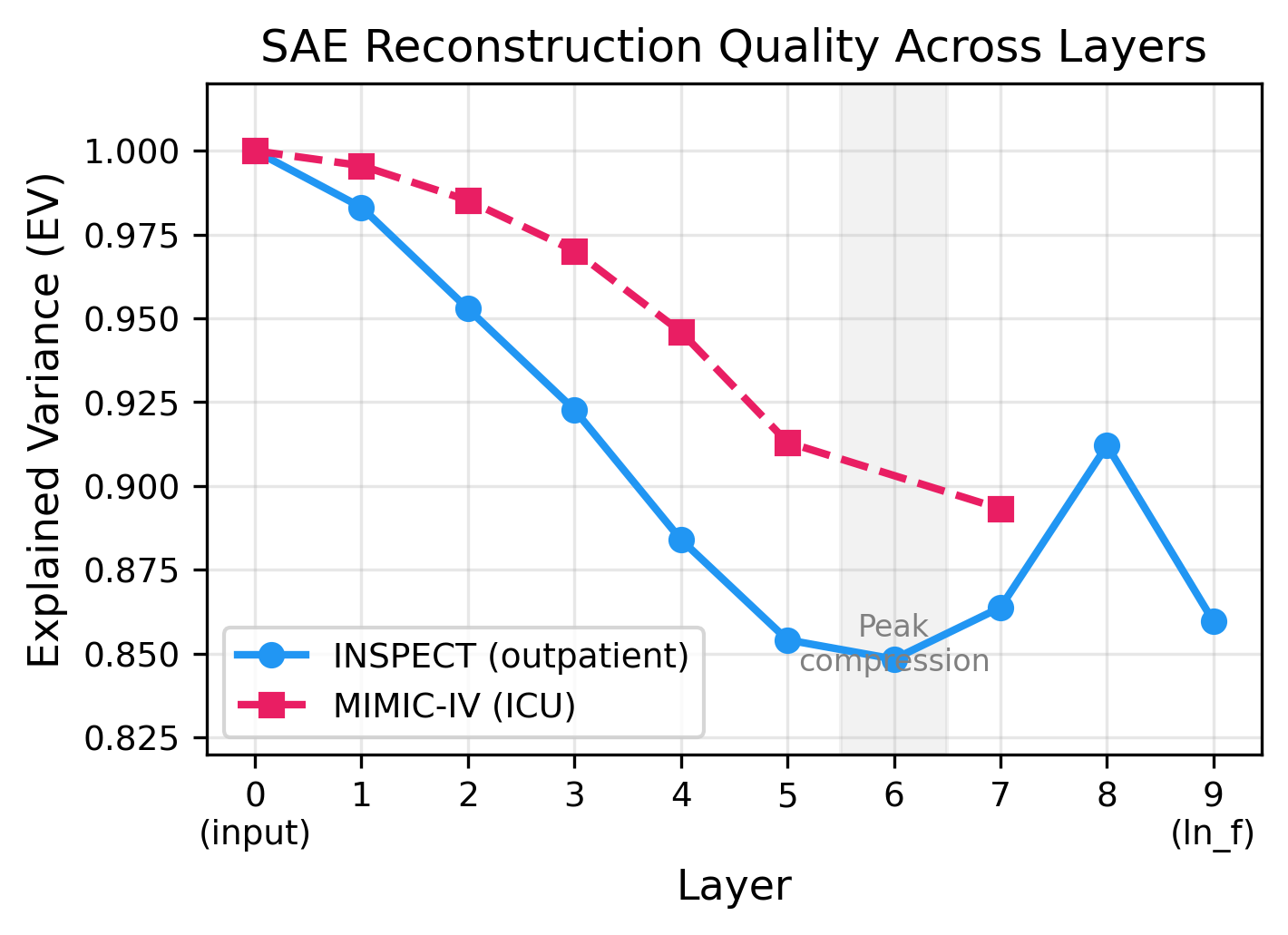}
\caption{Explained variance U-curve across transformer layers.  Sparse autoencoder reconstruction quality (explained variance) at each of 10 extraction points for INSPECT (outpatient) and MIMIC-IV (ICU).  Both datasets show a characteristic U-curve: near-perfect reconstruction at the input layer (layer 0), a trough at layer 6 corresponding to the most distributed representations, and partial recovery toward the output.}
\label{fig:ev-ucurve}
\end{figure}

Explained variance (EV) across layers followed a characteristic U-curve: near-perfect reconstruction at the input layer (EV $\approx$ 1.000), a trough at layer 6 (maximal sparse reconstruction difficulty), and partial recovery toward the output (Fig.~\ref{fig:ev-ucurve}).

\begin{table}[h]
\centering
\caption{Explained variance across selected layers (TopK SAE, expansion 8$\times$, $K$ = 16).  The U-curve is consistent across INSPECT and MIMIC-IV, with the trough at layer 6 corresponding to the most distributed representations under this SAE dictionary.  The development dataset sweep is shown in Supplementary Table~S2.}
\label{tab:ev-curve}
\small
\begin{tabular}{lcc}
\toprule
\textbf{Layer} & \textbf{INSPECT} & \textbf{MIMIC-IV} \\
\midrule
0 (input) & 1.000 & 1.000 \\
3 & 0.891 & 0.919 \\
6 (trough) & 0.848 & 0.895 \\
9 (post-ln\_f) & 0.860 & 0.884 \\
\bottomrule
\end{tabular}
\end{table}

The layer-6 trough indicates that these representations are the most difficult to reconstruct sparsely, consistent with distributed internal representations that are furthest from token-level identity.  The pattern replicated across INSPECT and MIMIC-IV (Table~\ref{tab:ev-curve}); the full 10-layer sweep including the development dataset is shown in Supplementary Table~S2.  eICU-CRD SAEs were trained for the outcome validation and mortality analysis (Sections~\ref{sec:specialisation}--\ref{sec:mortality}) but the full depth-wise complexity characterisation was not repeated on eICU.

\subsubsection{Feature complexity increases from token detectors to clinical concepts}

\begin{figure}[t]
\centering
\includegraphics[width=\textwidth]{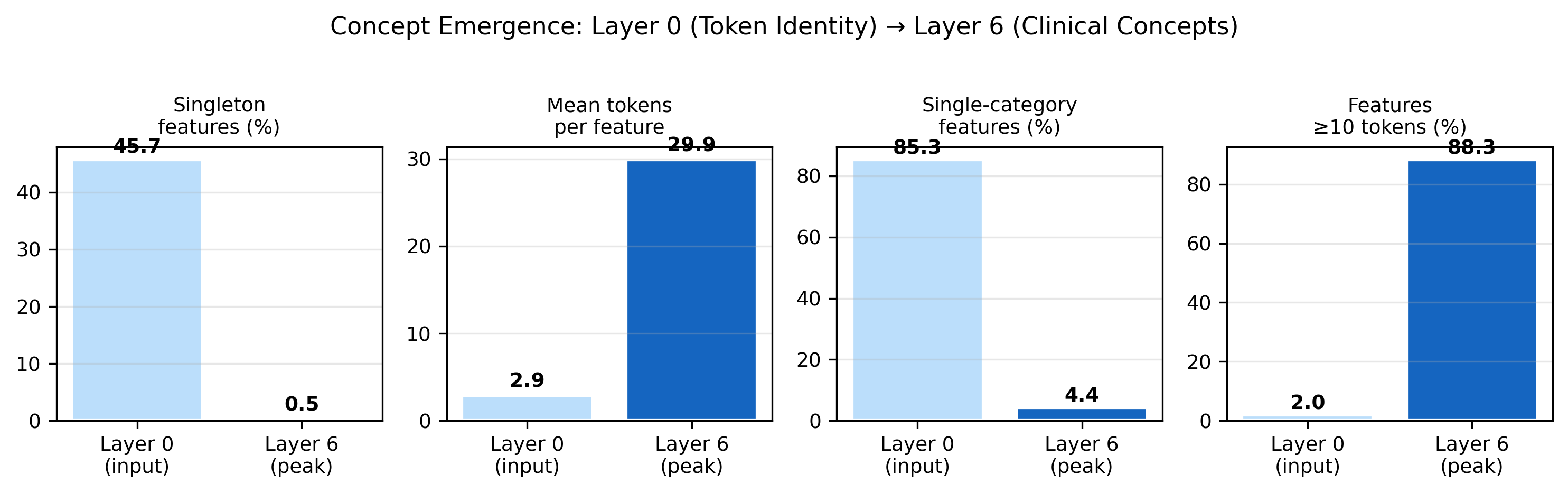}
\caption{Concept emergence from layer 0 to layer 6.  Four metrics quantifying the transition from token-level to concept-level representation in INSPECT SAEs.  Layer-0 features are near-perfect token detectors (45.7\% singleton, 2.9 mean tokens per feature, 85.3\% single-category).  Layer-6 features span approximately 30 tokens each across multiple clinical categories (0.5\% singleton, 29.9 mean tokens, 4.4\% single-category), representing a ten-fold increase in representational complexity.}
\label{fig:concept-emergence}
\end{figure}

Feature complexity increases ten-fold from layer 0 to layer 6, as quantified by complexity metrics across depth (\Cref{tab:concept-emergence}; Fig.~\ref{fig:concept-emergence}).

\begin{table}[h]
\centering
\caption{Feature complexity evolution (INSPECT SAEs).  By layer 6, features have progressed from near-perfect token detectors to distributed clinical concepts spanning multiple categories.}
\label{tab:concept-emergence}
\small
\begin{tabular}{lcccc}
\toprule
\textbf{Metric} & \textbf{Layer 0} & \textbf{Layer 6} \\
\midrule
Singleton features (\%) & 45.7 & 0.5 \\
Mean tokens per feature & 2.9 & 29.9 \\
Single-category features (\%) & 85.3 & 4.4 \\
Category entropy (nats) & 0.06 & 0.43 \\
\bottomrule
\end{tabular}
\end{table}

Layer-0 features were essentially one-to-one with input tokens: 45.7\% activated on a single token type, and 85.3\% drew activations from a single clinical category.  By layer 6, features had become distributed clinical concepts: each feature activated on a mean of 29.9 unique token types, only 4.4\% of features were confined to a single category, and category entropy had increased seven-fold (0.06 to 0.43 nats).  This ten-fold increase in mean tokens per feature quantifies the model's progressive abstraction from token identity to clinical concept.

At layer 6 on INSPECT, 92.2\% of features were category-coherent ($>$50\% of activations from a single clinical category), with 61.9\% strongly category-concentrated ($>$80\% single category).  This indicates that even at the layer of maximal sparse reconstruction difficulty, most features retain coherence within a clinical category while spanning multiple individual tokens within that category.  We note that category-level coherence is a weaker property than monosemanticity in the SAE literature, where the term typically implies interpretable single-concept features; our categories (LAB, MED, DX, etc.) are coarse groupings.

\subsubsection{Feature case studies}

We selected features with high category-concentration scores for illustrative case studies; given the 21\% cross-seed reproducibility (Section~\ref{sec:reproducibility}), individual feature identities should be interpreted with caution.

\paragraph{Feature 2187 (MIMIC-IV, layer 6): Elevated troponin with leukocytosis.}  Top token \tok{LAB:TROPONIN\_T:Q5} (high troponin, 30\% of activations), with secondary activations on \tok{LAB:WBC:Q5} (leukocytosis) and \tok{MED:ELECTROLYTE} (electrolyte replacement).  This feature was the single strongest mortality predictor by univariate Cox regression, illustrating that layer-6 features can capture clinically coherent multi-token patterns.

\paragraph{Feature 696 (MIMIC-IV, layer 6): ICU sedation pattern.}  \tok{MED:BENZO} (benzodiazepine) was the top token, with secondary activations on \tok{MED:OPIOID} and \tok{MED:PROPOFOL}---the combination characteristic of mechanically ventilated ICU patients requiring continuous sedation.  Its co-activation across three distinct drug classes demonstrates that layer-6 features encode treatment patterns rather than individual medications.

Three additional case studies (normal-range sodium, warfarin identity, leukocytosis with antibiotic use) are presented in Supplementary Material.

\subsubsection{Reproducibility}
\label{sec:reproducibility}

Feature reproducibility across three random seeds was 21\% (cosine similarity threshold 0.7), lower than the 30\% reported for large language model SAEs.\citep{paulo2025stability}  This likely reflects the smaller model scale (14.5M versus billions of parameters) and narrower vocabulary.  Individual features should therefore be interpreted as statistical properties of the learned representation, not as robust biomarkers.  The core qualitative findings (mortality prediction advantage at full sequence, task-specialisation pattern) appeared stable in the runs examined, but formal quantification of downstream seed-wise variability was not performed; the specific features driving aggregate performance varied across seeds.

Hyperparameter sweeps on INSPECT layer 6 (expansion $\in$ \{4, 8, 16\}, $K$ $\in$ \{8, 16, 32\}) confirmed that $K$ dominates explained variance while expansion has minimal effect beyond 4$\times$; the default configuration ($e$ = 8, $K$ = 16) sits at the Pareto knee of the sparsity--reconstruction trade-off.

\subsection{Task-dependent representational specialisation}
\label{sec:specialisation}

Across all datasets and all transformer layers tested, a consistent pattern was observed: SAE features outperformed dense layer representations for discrete clinical event prediction (mortality) while dense representations outperformed for continuous magnitude prediction (length-of-stay regression).  This comparison used simple mean-pooled logistic regression and ridge regression; the relative performance of sparse and dense representations may differ with more expressive downstream models.

The pattern was observed at layer 6 across all four datasets (\Cref{tab:specialisation-summary}).

\begin{table}[h]
\centering
\caption{Task-dependent specialisation at layer 6.  $\Delta$ = SAE minus dense representation performance.  Positive values indicate SAE superiority.  Full-sequence values used for consistency across datasets.}
\label{tab:specialisation-summary}
\small
\begin{tabular}{lcc}
\toprule
\textbf{Dataset} & \textbf{Mortality $\Delta$ AUC} & \textbf{LoS $\Delta R^2$} \\
\midrule
INSPECT & +0.118 & $-$0.540 \\
MIMIC-IV & +0.022 & $-$0.026 \\
eICU-CRD (zero-shot) & +0.031 & $-$0.025 \\
Dev (diabetes) & +0.075 & $-$0.908 \\
\bottomrule
\end{tabular}
\end{table}

The pattern is consistent with TopK sparsity ($K$ = 16) preserving the largest feature activations---corresponding to discrete clinical concepts such as ``elevated troponin'' or ``receiving vasopressors''---while discarding the small, distributed activations that encode continuous magnitudes.  This observation was made under one architecture, one extraction scheme, one pooling strategy, and simple downstream models; whether it generalises to other settings remains to be tested.

\subsubsection{$K$-sensitivity analysis}

\begin{figure}[t]
\centering
\includegraphics[width=0.8\textwidth]{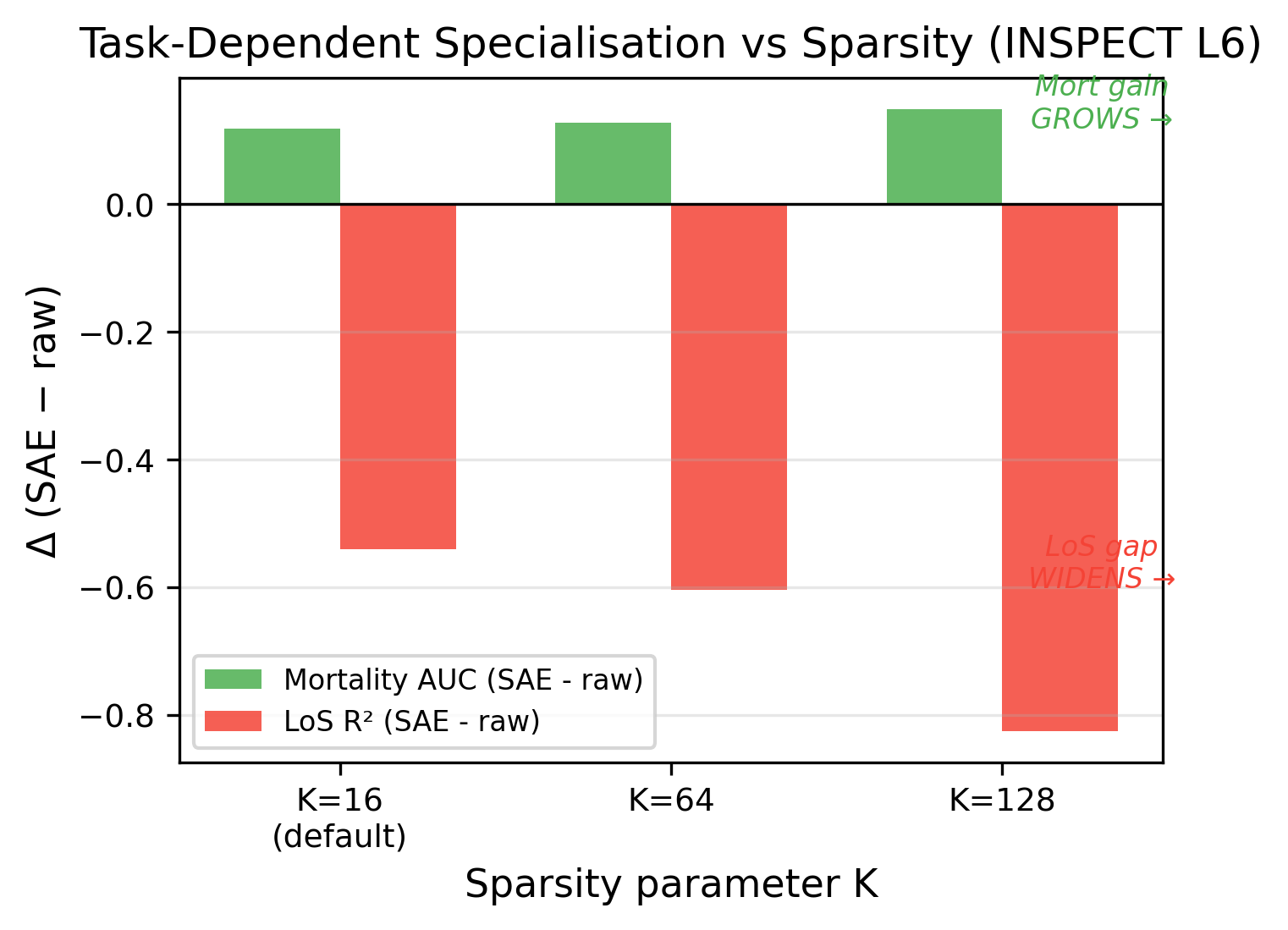}
\caption{Task-dependent specialisation with $K$-sensitivity analysis.  Mortality AUC (blue) and length-of-stay $R^2$ (red) for SAE features at three sparsity levels ($K$ = 16, 64, 128) versus $K$-independent dense layer representation performance (dashed lines).  INSPECT, layer 6, full-sequence.  The mortality advantage of SAE features increases with $K$ (0.929 to 0.960) while the length-of-stay disadvantage widens ($R^2$ from 0.133 to $-$0.152).}
\label{fig:k-sensitivity}
\end{figure}

To exclude the possibility that this pattern is an artefact of the sparsity threshold, we repeated the analysis at three levels of $K$ on INSPECT layer 6 (\Cref{tab:k-sensitivity}; Fig.~\ref{fig:k-sensitivity}).

\begin{table}[h]
\centering
\caption{$K$-sensitivity analysis (INSPECT, layer 6, full-sequence).  Mortality advantage of SAE features \emph{increases} with $K$, while length-of-stay disadvantage \emph{widens}.  Dense representation performance is $K$-independent.}
\label{tab:k-sensitivity}
\small
\begin{tabular}{lccc}
\toprule
\textbf{$K$} & \textbf{Mortality AUC (SAE)} & \textbf{LoS $R^2$ (SAE)} & \textbf{LoS $R^2$ (Dense)} \\
\midrule
16 & 0.929 & 0.133 & 0.673 \\
64 & 0.938 & 0.069 & 0.673 \\
128 & 0.960 & $-$0.152 & 0.673 \\
\bottomrule
\end{tabular}
\end{table}

The mortality advantage increased with higher $K$ (0.929 to 0.960), while the length-of-stay disadvantage widened ($R^2$ from 0.133 to $-$0.152).  If the pattern were a thresholding artefact, increasing $K$ should have narrowed both gaps by retaining more information from the original representation.  Instead, the divergence suggests task-dependent specialisation: sparse decomposition amplifies features relevant to discrete events at the expense of continuous magnitude information.  The pattern persists at $K$ = 128, which retains one-third of all features and is far from a harsh sparsity constraint.

Note that the $K$-sensitivity analysis uses full-sequence INSPECT data, not time-windowed data.  The task-dependent specialisation finding concerns a representational property of the SAE decomposition itself---the relationship between sparsity and task performance---not prediction at a specific clinically validated time window.

Under these full-sequence simple-probe conditions, the choice of representation depends on the prediction task: SAE features for discrete event prediction and dense layer representations for continuous magnitude prediction.  However, this pattern does not straightforwardly extend to leakage-safe clinically relevant windows, where dense representations are generally equal or superior for mortality prediction (Section~\ref{sec:mortality}).

\subsection{Leakage-safe mortality prediction}
\label{sec:mortality}

\subsubsection{The data leakage problem}

A critical concern for mortality prediction from clinical sequences is that events occurring near the end of a hospital stay---comfort care orders, withdrawal of treatment, terminal laboratory patterns---may encode the mortality outcome directly, inflating apparent predictive performance.  We developed a time-windowed evaluation methodology that restricts observations to clinically appropriate windows, transparently reporting both leakage-safe and contaminated results.

\subsubsection{Time-windowed analysis: ICU datasets}

\begin{figure}[t]
\centering
\includegraphics[width=\textwidth]{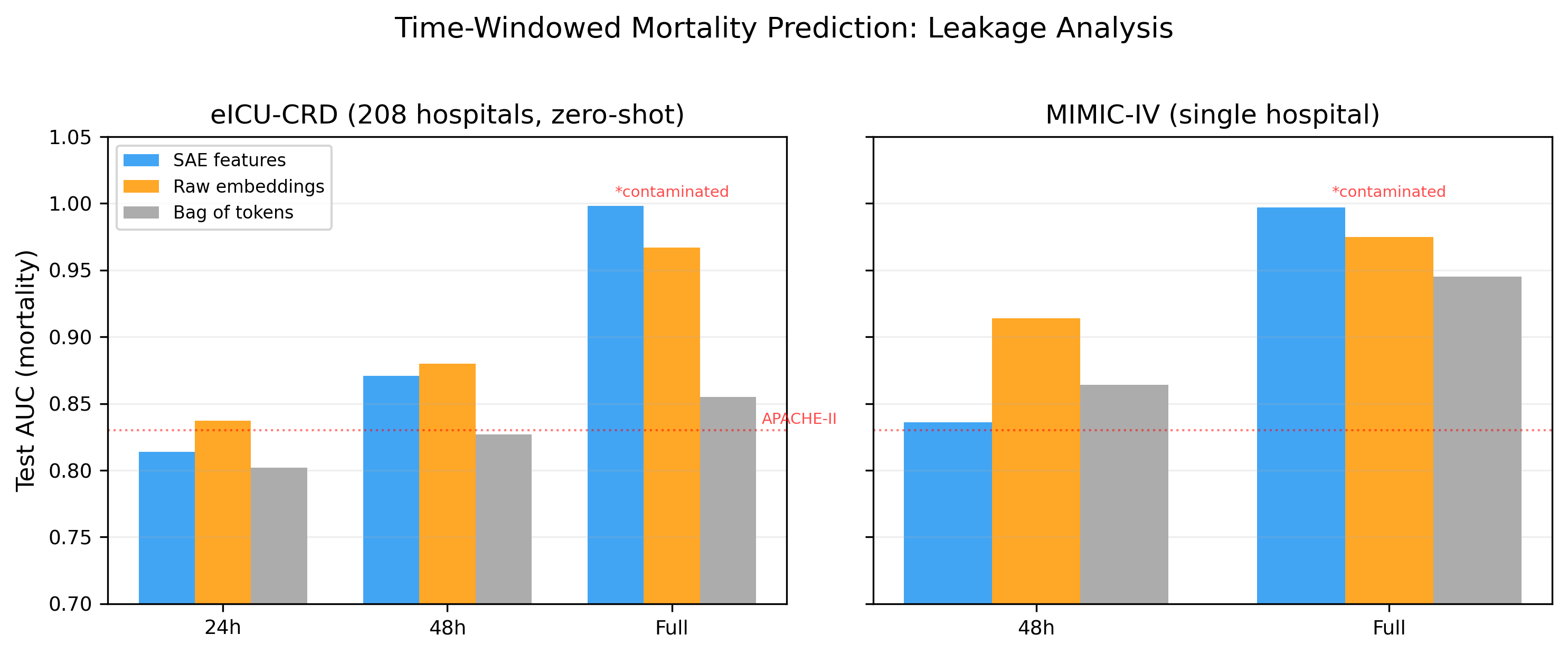}
\caption{Time-windowed mortality prediction across ICU and outpatient settings.  AUC for in-hospital mortality prediction at multiple observation windows on eICU-CRD (48h, full), MIMIC-IV (48h, full), and INSPECT (48h, 365d, 1y/3y, full).  Full-sequence AUCs are inflated by end-of-stay events that encode the outcome (marked with asterisks).  At leakage-safe windows, dense layer representations (orange) match or exceed SAE features (blue), with both outperforming bag-of-tokens baselines (grey) on most datasets.}
\label{fig:time-windowed}
\end{figure}

For ICU datasets (eICU-CRD and MIMIC-IV), we evaluated mortality prediction at three observation windows: 24 hours, 48 hours, and full sequence (\Cref{tab:time-windowed-icu}; Fig.~\ref{fig:time-windowed}).  The 48-hour window is the primary leakage-safe evaluation for ICU data.

\begin{table}[h]
\centering
\caption{Time-windowed mortality prediction on ICU datasets.  SAE = sparse autoencoder features (layer 6).  Dense = dense layer representations.  BoT = bag-of-tokens baseline ([DEATH] excluded).  eICU-CRD: base model zero-shot transferred from MIMIC-IV; SAE dictionary and downstream probes trained on eICU activations/data.  Bootstrap 95\% CIs shown for the primary eICU 48h comparison.  \textsuperscript{c}MIMIC-IV 48h CIs not available from stored results; the 0.078 gap between SAE and dense is large relative to typical bootstrap width at this sample size.  *Full-sequence results contaminated by end-of-stay information leakage---reported for transparency only.}
\label{tab:time-windowed-icu}
\small
\begin{tabular}{llccc}
\toprule
\textbf{Dataset} & \textbf{Window} & \textbf{SAE AUC [95\% CI]} & \textbf{Dense AUC [95\% CI]} & \textbf{BoT AUC} \\
\midrule
eICU-CRD & 48h & 0.871 [0.843, 0.903] & 0.880 [0.857, 0.903] & 0.827 \\
eICU-CRD & Full* & 0.998 & 0.967 & 0.855 \\
\midrule
MIMIC-IV & 48h & 0.836\textsuperscript{c} & 0.914\textsuperscript{c} & 0.864 \\
MIMIC-IV & Full* & 0.997 & 0.975 & 0.945 \\
\bottomrule
\end{tabular}
\end{table}

At the leakage-safe 48-hour window, SAE features and dense layer representations achieved broadly comparable mortality discrimination on eICU-CRD (0.871 versus 0.880), with both substantially outperforming the bag-of-tokens baseline (0.827).  On MIMIC-IV at 48 hours, dense representations outperformed SAE features (0.914 versus 0.836), and the bag-of-tokens baseline also outperformed SAE features (0.864 versus 0.836).

At leakage-safe observation windows, dense layer representations matched or outperformed sparse autoencoder features for mortality prediction across all ICU datasets tested.  On eICU-CRD, the cost of interpretability was small (SAE 0.871 versus dense 0.880, a difference of 0.009).  On MIMIC-IV at 48 hours, SAE features (0.836) were outperformed not only by dense representations (0.914) but also by the bag-of-tokens baseline (0.864).  This means that on MIMIC-IV, the SAE decomposition actively lost predictive information relative to a simple token-counting baseline---a result that should temper any claim that SAE features provide ``free'' interpretability.

\subsubsection{Time-windowed analysis: outpatient dataset}

For the outpatient INSPECT dataset, ICU-style 48-hour windows are inappropriate because deaths occur months to years after observation, not during active clinical deterioration.  We evaluated setting-appropriate observation windows (\Cref{tab:time-windowed-inspect}).

\begin{table}[h]
\centering
\caption{Mortality prediction on INSPECT (outpatient).  SAE = sparse autoencoder features (layer 6).  Dense = dense layer representations.  BoT = bag-of-tokens.  The 1-year observation / 3-year outcome window is the primary leakage-safe evaluation.  *Full-sequence result includes the complete clinical trajectory.}
\label{tab:time-windowed-inspect}
\small
\begin{tabular}{lccccr}
\toprule
\textbf{Window} & \textbf{SAE AUC} & \textbf{Dense AUC} & \textbf{BoT AUC} & \textbf{Deaths} \\
\midrule
\multicolumn{5}{l}{\textit{Primary leakage-safe evaluation}} \\
1y obs / 3y outcome & 0.697 & 0.800 & 0.785 & 41 \\
\midrule
\multicolumn{5}{l}{\textit{Underpowered exploratory windows}} \\
48h & 0.587 & 0.630 & 0.625 & $\sim$3 \\
365d (1 year) & 0.675 & 0.694 & 0.615 & $\sim$20 \\
Full* & 0.929 & 0.811 & 0.767 & 53 \\
\bottomrule
\end{tabular}
\begin{flushleft}
\footnotesize
Separately, an attention-pooled mortality head fine-tuned on the frozen base model achieved AUC 0.868 [0.822--0.916] at the 1-year/3-year window ($N$ = 1,906, 79 deaths on a different train/val/test split).  This is not directly comparable to the probe results above but demonstrates that task-specific fine-tuning improves substantially on mean-pooled linear readouts.
\end{flushleft}
\end{table}

At all leakage-safe windows on INSPECT, dense representations matched or exceeded SAE features for mortality prediction.  The 48-hour window has too few deaths ($\sim$3) for meaningful evaluation in an outpatient cohort.  At the 1-year observation / 3-year outcome window, dense representations outperformed SAE features (0.800 versus 0.697).  An attention-pooled mortality head fine-tuned on the frozen base model achieved AUC 0.868 [0.822--0.916] at this window (79 deaths, $N$ = 1,906), demonstrating that task-specific fine-tuning can improve performance beyond mean-pooled linear readouts.

Setting-dependent observation windows---48 hours for ICU data, 365 days or 1-year/3-year for outpatient data---provide an approach to leakage-safe evaluation, as most published mortality prediction results do not explicitly address the leakage risk posed by end-of-stay clinical events.

\subsection{Feature perturbation analysis}
\label{sec:perturbation}

To move beyond correlational evidence, we tested whether specific SAE features are associated with changes in the model's ability to generate treatment-appropriate clinical trajectories.  We selected the warfarin $\to$ INR association as the test case because it has the strongest directional signal in the companion paper (correct direction with the largest effect size among surrogate endpoints; DiD +0.052, Wilcoxon $p < 0.001$).

Four experiments targeted SAE features encoding high-INR concepts (identified by their top-token association with \tok{LAB:INR:Q4} and \tok{LAB:INR:Q5}).  Each experiment ablated targeted features---setting their SAE activations to zero via a forward hook---and re-ran the full warfarin $\to$ INR testing pipeline (200 patients, 200 samples per arm).  A novel delta-mode intervention was used to eliminate reconstruction noise (Methods).

The estimand is the change in the warfarin $\to$ INR difference-in-differences ($\Delta$DiD) caused by the intervention relative to the unperturbed model.  A negative $\Delta$DiD indicates that ablating or attenuating the targeted features reduced the warfarin-associated INR elevation---the expected direction if those features encode the drug--biomarker relationship.  A positive $\Delta$DiD indicates amplification.

\begin{table}[h]
\centering
\caption{Feature perturbation experiments on the warfarin $\to$ INR association.  $\Delta$DiD = change in warfarin effect size relative to unperturbed model; negative = attenuation (expected for ablation of INR-encoding features).  Targeted features encode high-INR concepts; random features are size-matched controls.  Ratio = $|$targeted$|$ / $|$random mean$|$.}
\label{tab:perturbation}
\small
\begin{tabular}{llccc}
\toprule
\textbf{Experiment} & \textbf{Mode} & \textbf{Targeted $\Delta$DiD} & \textbf{Random mean $\Delta$DiD} & \textbf{Ratio} \\
\midrule
Layer 6 reconstruct & Full SAE & +0.034 & +0.011 & 3.1$\times$ \\
Layer 6 delta & Delta-mode & $-$0.010 & +0.037 & opposite \\
Layer 9 delta & Delta-mode & $-$0.039 & $-$0.022 & 1.8$\times$ \\
Layer 9 attenuation $\times$5 & Scaled & $-$0.140 & $-$0.055 & 2.5$\times$ \\
\bottomrule
\end{tabular}
\end{table}

Across the four experiments, targeted INR features produced effects larger in magnitude than size-matched random feature sets in 3 of 4 conditions (1.8--3.1$\times$ the random mean); one condition produced an effect in the opposite direction to expectation.  The largest effect ($-$0.140 for layer-9 scaled latent attenuation) was 2.5$\times$ the random mean.  However, in no single experiment did the targeted effect fall strictly outside the distribution of random feature ablation effects.  These results are suggestive but not formally significant, reflecting the noise floor imposed by the SAE reconstruction quality (EV 0.85--0.90 at layer 6) and the limited number of random control sets (10--20 per experiment).

The delta-mode intervention is a methodological contribution regardless of the significance result.  By adding only the modification effect to the original hidden state rather than replacing it with the full SAE reconstruction, delta-mode reduced the noise floor by approximately 86$\times$ relative to na\"{i}ve reconstruct mode (from $\pm$0.43 to $\pm$0.005 in odds ratio units).

Retrospective demographic stratification of mortality prediction by race/ethnicity on eICU-CRD (48-hour window) is reported in Supplementary Table~S1.  Small event counts (28 Black deaths, 10 Hispanic deaths) preclude strong conclusions about equity; no formal calibration or equalized-odds analysis was performed.

% ═════════════════════════════════════════════════════════════════════════════
\section{Discussion}
\label{sec:discussion}
% ═════════════════════════════════════════════════════════════════════════════

\subsection{SAE decomposition of EHR model representations}

The SAE analysis provides a view of how FlatASCEND organises clinical knowledge internally.  The feature complexity pattern---from token detectors at layer 0 (45.7\% singleton) to distributed clinical concepts at layer 6 (0.5\% singleton, 29.9 mean tokens per feature)---quantifies the model's progressive abstraction.  The warfarin-encoding feature (Feature 60) that co-activates with INR measurements provides a concrete example: the same internal representation that encodes drug identity is plausibly involved in the warfarin $\to$ INR pharmacological association reported in the companion paper, though this link remains correlational rather than established by the perturbation analysis.

\subsection{Task-dependent specialisation}

The task-dependent specialisation finding is an observation that, to our knowledge, has not been previously reported for EHR models.  The pattern---SAE features outperforming for discrete events while dense layer representations outperform for continuous magnitudes---was consistent across datasets.  Most SAE evaluations assess features against a single downstream task.  By evaluating across multiple outcome types and confirming the pattern across three sparsity levels ($K$ = 16, 64, 128), we observe that representation choice interacts with task type under these specific experimental conditions.  The $K$-sensitivity analysis provides evidence against a thresholding artefact: if increasing $K$ narrowed both gaps, the pattern would be attributable to information loss; instead, the mortality advantage \emph{increased} with $K$ (0.929 to 0.960) while the length-of-stay disadvantage \emph{widened} ($R^2$ from 0.133 to $-$0.152).

This observation was made under one architecture, one extraction scheme, one pooling strategy, and very simple downstream models.  Whether this pattern generalises to other model architectures and more expressive readout methods remains to be established.  Concurrent work by Modi \etal\citep{modi2026medsae} found that SAE-based interventions are architecture-dependent in medical LLMs (effective for MedGemma, harmful for OpenBioLLM on the same retrieval task), suggesting that the interaction between sparse decomposition and downstream utility may be a general phenomenon across clinical AI modalities.

\subsection{Mortality prediction at leakage-safe windows}

At all leakage-safe observation windows, dense layer representations matched or exceeded SAE features for mortality prediction.  On MIMIC-IV at 48 hours, SAE features (0.836) were outperformed by both dense representations (0.914) and the bag-of-tokens baseline (0.864), meaning the SAE decomposition actively lost predictive information in that setting.  The best defensible claim is therefore narrow: SAE features preserve useful event-level signal on eICU-CRD zero-shot transfer (0.871 versus 0.880, a small gap) while enabling decomposition into interpretable units, but this does not generalise to all datasets.  The attention-pooled mortality head on INSPECT (AUC 0.868 at 1-year/3-year) demonstrates that task-specific fine-tuning can improve substantially on mean-pooled linear readouts, suggesting that the base representations are more informative than the simple evaluation pipeline reveals.

\subsection{Limitations}

\paragraph{SAE reproducibility.}  Feature stability across seeds is 21\%, meaning approximately four in five features are not stable across training runs.  This is lower than the 30\% reported for large language model SAEs\citep{paulo2025stability} and likely reflects the smaller model scale.  Individual features should be interpreted as statistical properties, not robust biomarkers.

\paragraph{Feature perturbation results are not formally significant.}  While targeted features produced effects 1.8--3.1$\times$ larger than random controls in 3 of 4 experiments, no experiment reached formal significance.  Larger random control sets and stronger SAEs (higher explained variance) would strengthen this analysis.

\paragraph{Simple downstream models.}  The mortality and length-of-stay comparisons use mean-pooled logistic/ridge regression.  More expressive downstream models (attention-pooled heads, fine-tuned classifiers) may change the relative performance of sparse versus dense representations.

\paragraph{No race/ethnicity tokens.}  The model has no explicit access to race or ethnicity information.  The fairness analysis is retrospective, and small event counts in minority subgroups preclude strong conclusions.

\paragraph{ICU-dominated open data.}  Two of three validation datasets are ICU cohorts.  Outpatient validation relies on a single dataset (INSPECT, 19,000 patients), limiting generalisability.

\paragraph{Single model architecture.}  All findings are from a single 14.5M-parameter model.  Whether SAE decomposition of larger EHR models produces similar patterns is unknown.

% ═════════════════════════════════════════════════════════════════════════════
\section{Methods}
\label{sec:methods}
% ═════════════════════════════════════════════════════════════════════════════

\subsection{FlatASCEND model}
\label{sec:architecture}

FlatASCEND is a 14.5-million-parameter GPT-2-style autoregressive transformer with ALiBi\citep{press2022alibi} positional encoding.  Each clinical event is represented as a single flat composite token (\eg, \tok{LAB:HBA1C:Q4}, \tok{MED:METFORMIN}, \tok{DX:SEPSIS}).  The model has dual output heads: a weight-tied content head for next-token prediction and a zero-inflated log-normal time head for continuous inter-event time prediction.  Laboratory and vital sign embeddings use factored representations (measure identity + ordinal level).  The scaled configuration uses 384-dimensional hidden states, 8 transformer layers, 12 attention heads, and 1536-dimensional feed-forward layers.

The model was trained on two open-access PhysioNet datasets (MIMIC-IV and INSPECT) and evaluated in zero-shot transfer on a third (eICU-CRD), plus a proprietary development dataset used for architecture design (\Cref{tab:datasets}).  The SAE analysis on MIMIC-IV and eICU-CRD used the drug-level 239-token vocabulary variant, which expands 6 medication classes to 33 individual drugs; this is why individual-drug tokens (e.g., \tok{MED:ANTICOAGULANT:WARFARIN}) appear in the feature case studies.  Generation quality and pharmacological association testing are reported in a companion paper;\citep{sainsbury2025flatascend} here we focus on the model's internal representations.

\begin{table}[h]
\centering
\caption{Dataset characteristics.  Dev = proprietary development dataset used for architecture design.  $\ddagger$The SAE and mortality analyses on MIMIC-IV used the drug-level 239-token vocabulary (6 medication classes expanded to 33 individual drugs), which is why individual-drug tokens appear in the feature case studies; the companion paper's generation and pharmacological testing used the 220-token class-level variant.  $\dagger$eICU-CRD uses the same 239-token vocabulary with MIMIC-derived quintile boundaries; the base model is zero-shot transfer (no adaptation), but SAE dictionaries were trained on eICU activations from this transferred model.}
\label{tab:datasets}
\small
\begin{tabular}{llcccrl}
\toprule
\textbf{Dataset} & \textbf{Setting} & \textbf{Hospitals} & \textbf{N} & \textbf{Vocabulary} & \textbf{Mortality} & \textbf{Access} \\
\midrule
Dev (diabetes) & Outpatient & 1 system & 61K patients & 76 tokens & --- & Proprietary \\
INSPECT & Outpatient & 1 & 19K patients & 220 tokens & 2.8\% & PhysioNet \\
MIMIC-IV & ICU & 1 & 425K admissions & 239 tokens\textsuperscript{$\ddagger$} & 2.6\% & PhysioNet \\
eICU-CRD & ICU & 208 & 200K stays & 239 tokens\textsuperscript{$\dagger$} & 5.3\% & PhysioNet \\
\bottomrule
\end{tabular}
\end{table}

\subsection{Sparse autoencoder training}
\label{sec:sae-methods}

\subsubsection{Architecture}

TopK SAE\citep{gao2024topk} with expansion factor 8$\times$ (384 $\to$ 3,072 features) and $K$ = 16 active features per input.  The encoder projects $\mathbf{h} \in \mathbb{R}^{384}$ to $\mathbf{z} = \text{TopK}(W_{\text{enc}} \mathbf{h} + \mathbf{b}_{\text{enc}}, K)$, retaining the $K$ largest activations.  The decoder reconstructs $\hat{\mathbf{h}} = W_{\text{dec}} \mathbf{z} + \mathbf{b}_{\text{dec}}$.  Decoder columns were normalised to unit length after each optimiser step.

\subsubsection{Streaming extraction}

Activations were extracted from the residual stream using PyTorch forward hooks at each extraction point (post-block layer normalisation).  A single forward pass through FlatASCEND produced activations at all 10 extraction points simultaneously.  Activations were consumed directly by the SAE training loop without disk materialisation.

\subsubsection{Dead feature resampling}

Following the methodology of Bricken \etal,\citep{bricken2023monosemanticity} features with zero activation across a 1,000-step sliding window were reinitialised by resampling from the highest-loss inputs in the current batch.  Resampling was checked every 500 steps.

\subsubsection{Training hyperparameters}

SAEs were trained with Adam (learning rate $3 \times 10^{-4}$, $\beta_1 = 0.9$, $\beta_2 = 0.999$) for 10,000 steps with batch size 16 patients per step.  Training loss was mean squared reconstruction error: $\mathcal{L}_{\text{SAE}} = \| \mathbf{h} - \hat{\mathbf{h}} \|_2^2$.  SAEs were trained independently at all 10 extraction points per dataset.  For eICU-CRD, the base FlatASCEND model is a zero-shot transfer from MIMIC-IV (no adaptation), but SAEs were trained on eICU activations from this transferred model.  The ``zero-shot'' designation thus applies to the base model, not to the SAE dictionary.

\subsection{Outcome validation}
\label{sec:outcome-methods}

\subsubsection{Per-patient mean feature activations}

For each patient: (1) forward pass through FlatASCEND on the full token sequence (or windowed subsequence for time-windowed analysis); (2) extract residual stream activation at the target layer; (3) encode through the trained SAE to obtain a 3,072-dimensional sparse feature vector per token position; (4) compute the mean across all token positions, yielding a single feature vector per patient.  The same procedure was applied to dense layer representations (384-dimensional mean across positions) without SAE encoding.

\subsubsection{Mortality prediction}

An L2-regularised logistic regression classifier was trained on per-patient feature vectors with the regularisation parameter $C$ selected by 5-fold cross-validation ($C \in \{0.001, 0.01, 0.1, 1, 10, 100\}$).  AUC-ROC was the primary metric.  Bootstrap 95\% confidence intervals were computed from 500 resamples.

\subsubsection{Clinical baselines}

Three layer-independent baselines: (1) bag of tokens (normalised token counts, \tok{[DEATH]} excluded); (2) token presence (binary indicators per token type); (3) sequence length (single feature).

\subsubsection{Time-windowed analysis}

For the time-windowed evaluation, each patient's token sequence was truncated to include only events from the specified observation window.  Observation windows were defined using cumulative inter-event times from the source data (not model-predicted times).  For each patient, the cumulative sum of recorded time deltas was compared against the specified window; tokens beyond this threshold were excluded from feature computation but retained for outcome ascertainment.  For ICU datasets, 24-hour and 48-hour windows were used; for outpatient data (INSPECT), 48-hour, 365-day, and 1-year observation / 3-year outcome windows were evaluated.

\subsubsection{Demographic extraction}

Sex and age decile were extracted from demographic tokens in the first 20 positions of each patient's sequence.  Race and ethnicity for eICU-CRD were extracted from patient demographic tables (not from model tokens, which do not include race/ethnicity).

\subsubsection{Survival analysis}

Univariate Cox proportional hazards regression was fitted per SAE feature with Bonferroni correction ($\alpha = 0.05 / 3072$).  Harrell's C-index was computed using concordant/discordant pair counting.

\subsection{Delta-mode feature perturbation}
\label{sec:delta-mode}

The na\"{i}ve approach to SAE-based feature perturbation replaces the hidden state entirely with the SAE reconstruction, introducing approximately 10\% reconstruction noise at every position.  For single-feature interventions, this noise dominates the signal.

Our delta-mode intervention adds only the modification effect to the original hidden state:
\begin{equation}
    \hat{\mathbf{h}} = \mathbf{h} + \left( W_{\text{dec}} \cdot (\mathbf{z} \odot \mathbf{m}) + \mathbf{b}_{\text{dec}} \right) - \left( W_{\text{dec}} \cdot \mathbf{z} + \mathbf{b}_{\text{dec}} \right)
\end{equation}
where $\mathbf{m}$ is a mask vector (0 for ablated features, 1 otherwise).  When no features are modified ($\mathbf{m} = \mathbf{1}$), this reduces exactly to identity ($\hat{\mathbf{h}} = \mathbf{h}$), eliminating the noise floor entirely.  This reduced the noise floor by approximately 86$\times$ relative to the reconstruct mode (from $\pm$0.43 to $\pm$0.005 in odds ratio units).

The scaled latent attenuation variant used $\mathbf{m}_i = 1 - \alpha \cdot \sigma_i$ for targeted features, where $\alpha$ controls the attenuation magnitude.  Although the modification is multiplicative in the latent feature space, the resulting hidden-state intervention is additive via the delta-mode construction.  Random feature ablation controls used the same number of features selected uniformly at random (10--20 random sets per experiment).

\subsection{Statistical analysis}

Mortality AUC confidence intervals were computed by bootstrap resampling (500 iterations).  Task-dependent specialisation was assessed at three sparsity levels ($K$ = 16, 64, 128) as a sensitivity analysis.  Feature reproducibility was measured by pairwise cosine similarity of decoder weight columns across three random seeds, with features matched using the Hungarian algorithm; 21\% of features exceeded a cosine similarity of 0.7 across all seed pairs.  All $p$-values are two-sided.

% ═════════════════════════════════════════════════════════════════════════════
\section*{AI Technology Disclosure}
% ═════════════════════════════════════════════════════════════════════════════

Claude (Anthropic) was used for analysis assistance, code review, and manuscript editing during the preparation of this work.  All scientific decisions, experimental design, data analysis, and interpretation were performed by the authors.

% ═════════════════════════════════════════════════════════════════════════════
\section*{Ethics and Data Access}
% ═════════════════════════════════════════════════════════════════════════════

Development dataset access was granted under an institutional data governance framework for research use.  MIMIC-IV data was accessed through PhysioNet under the PhysioNet Credentialed Health Data Use Agreement.  INSPECT is available through PhysioNet under the Stanford University data use agreement.  eICU-CRD data was accessed through PhysioNet.  All analyses were conducted on de-identified data.

% ═════════════════════════════════════════════════════════════════════════════
\section*{Code and Data Availability}
% ═════════════════════════════════════════════════════════════════════════════

The model architecture, training pipeline, sparse autoencoder infrastructure, and outcome validation code are hosted at \url{https://github.com/csainsbury/ascend-flat} (currently private; contact the corresponding author for collaborator access).  MIMIC-IV, INSPECT, and eICU-CRD data are available through PhysioNet\citep{goldberger2000physionet} with credentialed access.

\clearpage
% ═════════════════════════════════════════════════════════════════════════════
\section*{Supplementary Material}
% ═════════════════════════════════════════════════════════════════════════════

\subsection*{Supplementary Table S1: Demographic stratification of mortality prediction (eICU-CRD)}

\begin{table}[H]
\centering
\small
\begin{tabular}{lrrccc}
\toprule
\textbf{Group} & \textbf{N} & \textbf{Deaths} & \textbf{SAE AUC} & \textbf{Dense AUC} & \textbf{BoT AUC} \\
\midrule
Overall & 5,000 & 266 & 0.871 & 0.880 & 0.827 \\
\midrule
White & 3,811 & 206 & 0.864 & 0.875 & 0.830 \\
Black & 568 & 28 & 0.911 & 0.913 & 0.838 \\
Hispanic & 210 & 10 & 0.878 & 0.861 & 0.853 \\
Other/Unknown & 354 & 19 & 0.878 & 0.881 & 0.786 \\
\bottomrule
\end{tabular}
\begin{flushleft}
\footnotesize
48-hour observation window, eICU-CRD zero-shot transfer, layer 6.  Small event counts in minority subgroups (28 Black deaths, 10 Hispanic deaths) limit the precision of these comparisons and preclude strong conclusions about equity.  No formal calibration or equalized-odds analysis was performed.  Race/ethnicity extracted from eICU patient demographic tables (not from model tokens).
\end{flushleft}
\end{table}

\subsection*{Supplementary Table S2: Full 10-layer explained variance sweep}

\begin{table}[H]
\centering
\small
\begin{tabular}{lcc}
\toprule
\textbf{Layer} & \textbf{INSPECT} & \textbf{MIMIC-IV} \\
\midrule
0 (input) & 1.000 & 1.000 \\
1 & 0.972 & 0.984 \\
2 & 0.927 & 0.952 \\
3 & 0.891 & 0.919 \\
4 & 0.874 & 0.910 \\
5 & 0.859 & 0.901 \\
6 (trough) & 0.848 & 0.895 \\
7 & 0.855 & 0.891 \\
8 & 0.857 & 0.886 \\
9 (post-ln\_f) & 0.860 & 0.884 \\
\bottomrule
\end{tabular}
\begin{flushleft}
\footnotesize
TopK SAE, expansion 8$\times$, $K$ = 16.  The U-curve shape is consistent across both datasets, with the trough at layer 6 corresponding to the most distributed representations.  Intermediate layers (1--5) show monotonic decline from input.
\end{flushleft}
\end{table}

\subsection*{Supplementary Table S3: Hyperparameter sensitivity analysis (INSPECT, layer 6)}

\begin{table}[H]
\centering
\small
\begin{tabular}{ccccc}
\toprule
\textbf{Expansion} & \textbf{$K$} & \textbf{EV} & \textbf{Mortality AUC} & \textbf{Features} \\
\midrule
4 & 8 & 0.669 & 0.928 & 1,536 \\
4 & 16 & 0.842 & 0.929 & 1,536 \\
4 & 32 & 0.932 & 0.933 & 1,536 \\
8 & 8 & 0.671 & 0.916 & 3,072 \\
\textbf{8} & \textbf{16} & \textbf{0.848} & \textbf{0.929} & \textbf{3,072} \\
8 & 32 & 0.939 & 0.938 & 3,072 \\
16 & 8 & 0.672 & 0.920 & 6,144 \\
16 & 16 & 0.851 & 0.932 & 6,144 \\
16 & 32 & 0.941 & 0.940 & 6,144 \\
\bottomrule
\end{tabular}
\begin{flushleft}
\footnotesize
$K$ dominates explained variance (EV) while expansion factor has minimal effect beyond 4$\times$.  The default configuration ($e$ = 8, $K$ = 16, bold) sits at the Pareto knee of the sparsity--reconstruction trade-off.  Mortality AUC is robust across configurations (range 0.916--0.940).
\end{flushleft}
\end{table}

\subsection*{Supplementary Table S4: Feature perturbation experiments (full detail)}

\begin{table}[H]
\centering
\small
\begin{tabular}{llccccl}
\toprule
\textbf{Experiment} & \textbf{Mode} & \textbf{Targeted $\Delta$} & \textbf{Random mean} & \textbf{Random range} & \textbf{Ratio} & \textbf{Significant?} \\
\midrule
Layer 6 reconstruct & Full SAE & +0.034 & +0.011 & [$-$0.09, +0.09] & 3.1$\times$ & No \\
Layer 6 delta & Delta & $-$0.010 & +0.037 & [$-$0.01, +0.04] & opposite & No \\
Layer 9 delta & Delta & $-$0.039 & $-$0.022 & [$-$0.06, +0.01] & 1.8$\times$ & No \\
Layer 9 attenuation $\times$5 & Scaled & $-$0.140 & $-$0.055 & [$-$0.14, +0.02] & 2.5$\times$ & No \\
\bottomrule
\end{tabular}
\begin{flushleft}
\footnotesize
Warfarin $\to$ INR association, 200 patients, 200 samples per arm.  Targeted features encode high-INR concepts (\tok{LAB:INR:Q4}, \tok{LAB:INR:Q5}).  Random controls: 10--20 size-matched random feature sets per experiment.  Targeted effects are larger in magnitude than random mean in 3/4 conditions (one condition opposite to expected direction) but do not fall outside the random distribution in any single experiment.
\end{flushleft}
\end{table}

\subsection*{Supplementary Table S5: Cross-seed feature reproducibility (INSPECT)}

Feature stability was measured by pairwise cosine similarity of decoder weight columns across 3 random seeds, matched using the Hungarian algorithm.  At cosine threshold 0.7:

\begin{table}[H]
\centering
\small
\begin{tabular}{lcc}
\toprule
\textbf{Layer} & \textbf{Matched (\%)} & \textbf{Mean cosine (matched)} \\
\midrule
0 (input) & 42.3 & 0.91 \\
3 & 24.1 & 0.82 \\
6 & 21.0 & 0.79 \\
9 & 19.8 & 0.78 \\
\bottomrule
\end{tabular}
\begin{flushleft}
\footnotesize
Reproducibility decreases with depth, consistent with deeper features being more distributed and less tied to specific input tokens.  The 21\% figure cited in the main text is for layer 6.  Layer-0 features (42\% matched) are more stable because they are near-token detectors.
\end{flushleft}
\end{table}

\subsection*{Supplementary: Additional feature case studies}

\paragraph{Feature 658 (MIMIC-IV, layer 6): Leukocytosis with antibiotic use.}  Dominated by \tok{LAB:WBC:Q5}, this feature activated on 32 unique token types spanning laboratory and medication categories, capturing the co-occurrence of leukocytosis with antibiotic use, fluid administration, and vasopressor support.

\paragraph{Feature 31 (INSPECT, layer 6): Normal-range sodium.}  \tok{LAB:SODIUM:Q3} (normal sodium, quintile 3) accounted for the dominant activation with 100\% category purity.  This illustrates that even at the most distributed layer, some features remain tied to single laboratory values rather than becoming multi-concept abstractions.

\paragraph{Feature 60 (MIMIC-IV, layer 6): Warfarin identity.}  \tok{MED:ANTICOAGULANT:WARFARIN} dominated (84\% top-1 fraction).  The residual 16\% of activations on related tokens (other anticoagulants, INR measurements) suggests this feature encodes warfarin use as a clinical concept.  The perturbation analysis did not establish a direct causal link from this specific feature to the warfarin $\to$ INR pharmacological association reported in the companion paper.

% ═════════════════════════════════════════════════════════════════════════════
% ── Bibliography ──────────────────────────────────────────────────────────
% ═════════════════════════════════════════════════════════════════════════════

\bibliographystyle{plainnat}

\end{document}